%% file: main.tex
\documentclass[conference]{ieeeconf}
\usepackage{cite}
\usepackage{amsmath,amssymb,amsfonts}
\usepackage{algorithm}
\usepackage{algorithmic}
\usepackage{graphicx}
\usepackage{textcomp}
\usepackage{xcolor}
\usepackage{array}
\usepackage{subcaption}
\usepackage{makecell}
\usepackage{booktabs}
\usepackage{wasysym}
\usepackage{placeins}
\usepackage[utf8]{inputenc}
\usepackage{caption}
\usepackage{xurl}

\makeatletter

\def\section{\@startsection{section}{1}{\z@}{1.0ex plus 0.5ex minus 0.5ex}%
{0.4ex plus 0.2ex minus 0ex}{\normalfont\normalsize\centering\scshape}}
\def\subsection{\@startsection{subsection}{2}{\z@}{0.9ex plus 0.5ex minus 0.4ex}%
{0.3ex plus 0.2ex minus 0ex}{\normalfont\normalsize\itshape}}
\makeatother

\begin{document}

\title{\textbf{Suspended-Fabric Robotic Manipulation Surfaces: Modelling, Design Space, and Multimodular Extension}}

\author{Pratik Ingle, J{\o}rn Lambertsen, Kasper St{\o}y, Andres Fai\~na\\
\small IT University of Copenhagen, Denmark\\
\small \{prin, jrnl, ksty, anfv\}@itu.dk}

\maketitle

\input{sections/abstract}

\input{sections/introduction}

\input{sections/platform}
\input{sections/catenary_model}

\input{sections/model_validation}

\input{sections/control}
\input{sections/design}
\input{sections/multimodular}
\input{sections/discussion}
\input{sections/conclusion}

\bibliographystyle{unsrt}
\bibliography{ref}

\end{document}

%% file: sections/abstract.tex
\begin{abstract}

Robotic Manipulation Surfaces (RMS) reposition objects by deforming the surface on which they rest. A recent class of RMS suspends a continuous fabric sheet from a sparse
grid of vertical linear actuators lowering the actuator density of such surfaces but how to design such surfaces so that objects can be controlled everywhere on it, and how to extend it beyond a single module so that objects can pass between modules and achieve parallel, remain open. This paper presents a separable catenary model of the suspended sheet that is fast to compute and provides both the surface shape for given actuator heights and the actuator heights for a desired local slope. The closed form inverse lets a simple PID controller position objects on the sheet, and the model relates fabric length, module span and actuator stroke to the reachable workspace, yielding a minimum fabric length bound and design charts for these surfaces. The approach is then extended to a multimodular platform in which neighbouring modules share actuators. On a $2 \times 2$ prototype, we demonstrate that centimetre-scale objects are transferred between modules and positioned within 0.03 m over a $1 m^2$ workspace with only nine actuators, enabling multi-object manipulation and opening the way to parallel manipulation.
\end{abstract}


%% file: sections/introduction.tex
\section{Introduction}

\begin{figure*}[t]
  \centering
  \includegraphics[width=0.75\linewidth]{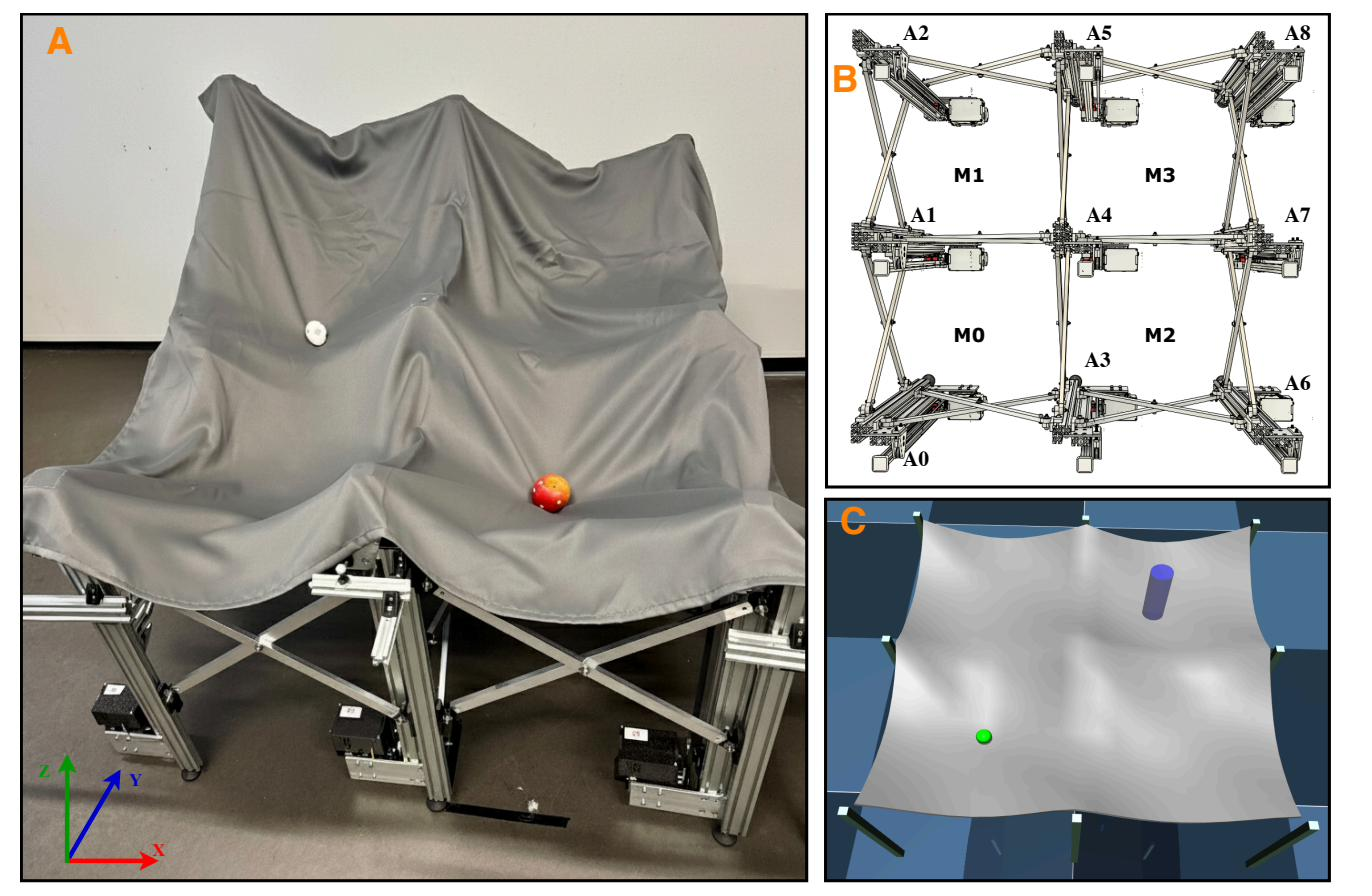}
  \caption{Multimodular soft manipulation surface. (A) Physical platform
  during manipulation of a spherical object. (B) Top view of the nine
  actuators (A0 to A8) and the four modules (M0 to M3) with the fabric
  removed. Adjacent modules share two actuators and diagonal modules share
  one. (C) MuJoCo model of the $2 \times 2$ configuration showing an
  object (green sphere) and its target location (blue cylinder).}
  \label{fig:platform}
\end{figure*}

Food processing, logistics and warehouse automation require large numbers
of heterogeneous and often fragile objects to be repositioned over
extended work areas. Pick and place manipulators handle such objects
sequentially, which limits throughput and risks damaging delicate items,
while conveyor belts provide throughput but only unidirectional transport.
Robotic Manipulation Surfaces (RMS) follow a different strategy: rather
than grasping an object, they deform or actuate the surface on which the
object rests, and the resulting distributed contact allows safe and
parallel handling of diverse items without object specific end effectors.
Established RMS designs, such as arrays of
rollers~\cite{parajuli2014actuator}, vertical
pistons~\cite{xue2024arraybot} or linear delta robots~\cite{Patil_2023},
obtain fine control through high actuator density: ArrayBot drives 256
pistons at a 20\,mm pitch, about 2500 actuators per square metre, and
the delta array of~\cite{Patil_2023} actuates 192 degrees of freedom
over an $8 \times 8$ grid. Such density limits the workspace that can be
built and maintained in practice.

Some RMS decouple control resolution from actuator density by covering
a sparse set of actuators with a continuous elastic layer that deforms
between them. Festo's wave handling surface~\cite{festo} placed such a
layer over a grid of pneumatic actuators and Morpho~\cite{yu2008morpho}
used membrane deformation to transfer objects between modules, both
demonstrated with rolling objects only. A fabric sheet suspended from
four vertical linear actuators at the corners of a $0.5$\,m square
module (16 actuators$/m^2$) has been shown to manipulate
fragile and irregular objects at an object to actuator size ratio as
low as 0.01~\cite{11020841}, using a geometric transformation combined
with a PID controller~\cite{11163787}.

A parallel line of work manipulates objects on a deformable sheet whose
supports are themselves mobile. Early work repositioned a rigid body on a
flexible sheet for an automated rehabilitation
bed~\cite{roy2005repositioning}. Subsequently, a team of mobile robots
holding a sheet has been used to transport and manipulate carried objects,
described either through the strain energy of an elastic
membrane~\cite{HUNTE2021437} or through the virtual variable cables model
(VVCM), which replaces the sheet by a set of dynamically changing
cables~\cite{hu2022multirobot} and has been extended to stable kinematics
for arbitrary team formations~\cite{11353119}. Both descriptions serve as
baselines in Section~\ref{sec:validation}. Hanging cloth itself has been
modelled by fitting catenary curves between the constraint points and
relaxing the result~\cite{weil1986synthesis}, whereas membrane models
that resolve the taut, wrinkled and slack states of a sheet require
numerical solution~\cite{zhang2024wrinkling}.

In summary, existing deformable sheet and cloth
models~\cite{roy2005repositioning, HUNTE2021437, hu2022multirobot,
11353119, weil1986synthesis, zhang2024wrinkling} give the shape of a
sheet for given support positions, but none maps a desired local slope
at the object directly to the heights of four vertical supports in the
closed form that an online controller requires. In addition, the design
of this type of RMS is challenging, as no analytical link relates fabric
slack, module size and actuator stroke to the workspace over which an
object can be held, so these parameters have been chosen by trial and
error. Finally, extending the surface beyond one module without a high
actuator count requires neighbouring modules to share actuators, which
couples them mechanically, and no route to large area scaling under that coupling has been shown.

This paper expands previous work on suspended fabric RMS and provides
three contributions toward extensible, large area object manipulation
with sparse actuation.

\emph{A simplified fabric model with an analytical inverse.} The fabric
is modelled as a separable catenary whose single sag parameter is fixed
by fabric length and module span, giving the surface shape for given
actuator heights and the actuator heights for a desired local slope in
closed form. The low computational cost and the inverse kinematics of
the model allow it to be used in online controllers, which improves the available workspace of the system.
 
\emph{Design conditions for suspended fabric surfaces.} Because the
model depends on fabric slack, module span and actuator stroke, the
trade off between these three parameters is analysed in closed form,
yielding a minimum fabric length bound and reachable workspace charts
from which the conditions that maximise the workspace are read directly.
 
\emph{Extension to a multimodular surface.} The suspended fabric
approach is extended to a multimodular prototype with shared actuators, demonstrating transfer and positioning of heterogeneous objects across modules while characterising the inter module coupling that actuator sharing introduces. Such a surface could, for example, be spread under an orchard tree to catch shaken fruit and convey it undamaged to one module for packing.

%% file: sections/platform.tex
\section{Platform and Simulation}
\label{sec:platform}

\subsection{Hardware}


The prototype (Fig.~\ref{fig:platform}) consists of nine linear actuators on a scissor frame that sets their spacing; at the $0.5$\,m spacing used here they form a $2 \times 2$ grid of modules of span $D = 0.5$\,m (Section~\ref{sec:multimodular}) resting at $0.7$\,m. Each actuator is stepper driven with 12 bit encoder feedback, has a stroke of 0.4\,m, and is commanded within $\pm u_{\max} = 0.16$\,m of the resting height throughout. The surface is a $1.2 \times 1.2$\,m polyester sheet, giving $S = 0.6$\,m of material across each $0.5$\,m module span. Objects are tracked by a six camera OptiTrack system at 120\,Hz.

\subsection{Simulation Environment}


MuJoCo~\cite{todorov2012mujoco} is used for simulation. The fabric is a \texttt{flexcomp} mesh with the membrane elasticity plugin (Young modulus 3\,kPa, Poisson ratio 0.3, thickness 1\,mm), which resists stretching but not bending; the corner vertices are fixed to prismatic actuators and the solver uses an implicit fast integrator at 1\,ms. A single module is a $15 \times 15$ mesh at 35\,mm spacing, material covers the 0.5 m pitch, so the simulated sheet is in tension where the physical sheet is slack; the $2 \times 2$ layout of Fig.~\ref{fig:platform}C is a $21 \times 21$ mesh at 49\,mm and is used only for the end to end trials of Section~\ref{sec:target}. All other simulation results use the single module, whose fidelity to the physical fabric is assessed in Section~\ref{sec:validation}.

%% file: sections/catenary_model.tex
\section{Separable Catenary Surface Model}
\label{sec:model}

\begin{figure*}[t]
  \centering
  \includegraphics[width=\textwidth]{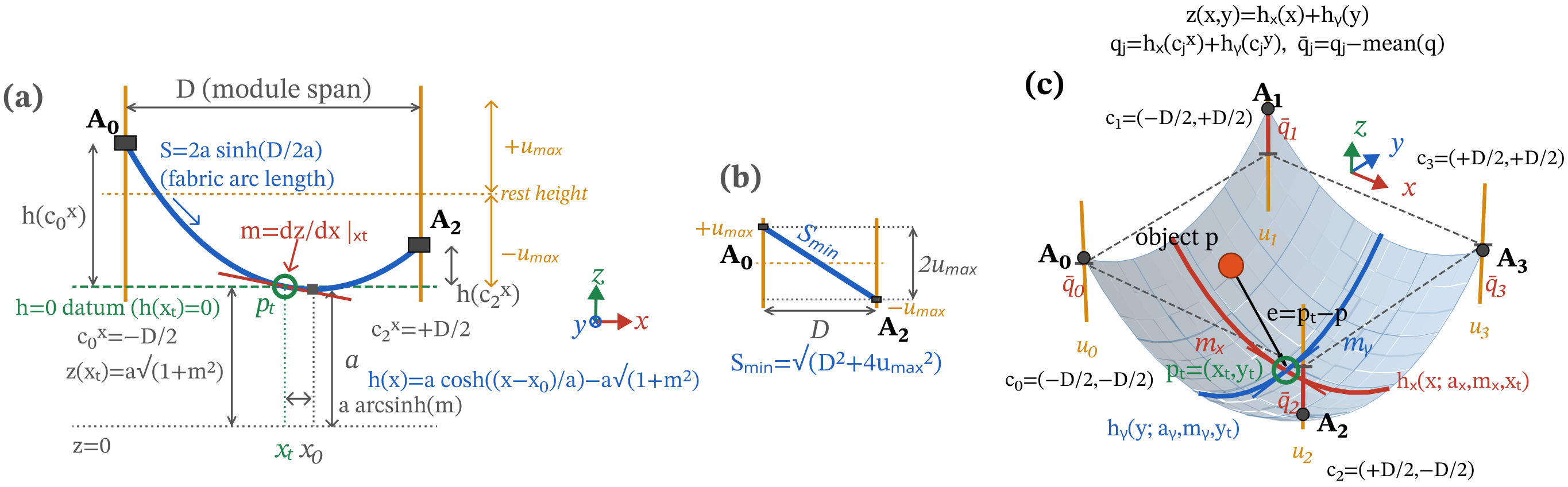}
  \caption{Separable catenary surface model on a single module.
  \textbf{(a)} Single axis catenary of \eqref{eq:catenary} to
  \eqref{eq:arc_length} between the supports of one module: the vertex at
  $x_0$ lies $a$ above the directrix, the commanded slope $m$ is the
  tangent at the target $x_t$, and the corner heights $h(c_j^x)$ are
  measured from the datum $h(x_t) = 0$. The fabric length $S$ and the
  module span $D$ fix $a$ through \eqref{eq:arc_length}.
  \textbf{(b)} Minimum fabric length of \eqref{eq:smin}.
  \textbf{(c)} Surface of \eqref{eq:separable} as the sum of the two cross
  sections through $\mathbf{p}_t = (x_t, y_t)$, with the corner heights
  $\bar{q}_j$ of \eqref{eq:mean_remove} measured from the mean plane
  (dashed) and the error vector $\mathbf{e}$ driving the controller of
  Section~\ref{sec:controller}.}
  \label{fig:catenary}
\end{figure*}


The model treats the fabric as a uniform and inextensible sheet that hangs
under its own weight between supports, so that each cross section adopts
the classical catenary profile. Three consequences follow. The profile is
governed by a single scalar sag parameter rather than by the elastic
constants of the material. Local indentation caused by the mass of the
carried object is not represented. Cross axis coupling is neglected,
since the two axes are treated as independent. The error/cost of these assumptions are measured empirically in section~\ref{sec:validation}. 


\subsection{Single Axis Catenary}

The cross sectional profile of a sheet suspended between two points is
\begin{equation}
  z(x) = a \cosh\!\left(\frac{x - x_0}{a}\right),
  \label{eq:catenary}
\end{equation}
where $a > 0$ is the sag parameter, with larger $a$ producing a flatter
profile, and $x_0$ is the horizontal position of the vertex. The slope is
\begin{equation}
  \frac{dz}{dx} = \sinh\!\left(\frac{x - x_0}{a}\right).
  \label{eq:slope}
\end{equation}
For the curve to be usable as a control model, the vertex must be placed
so that the slope at the target position $x_t$ equals a commanded value
$m$. Setting $dz/dx|_{x_t} = m$ in~\eqref{eq:slope} gives
\begin{equation}
  x_0 = x_t - a\,\mathrm{arcsinh}(m).
  \label{eq:vertex}
\end{equation}
The controller requires corner heights relative to the target, so
$h(x) = z(x) - z(x_t)$ is defined. Substituting~\eqref{eq:vertex}
into~\eqref{eq:catenary} and using
$\cosh(\mathrm{arcsinh}(m)) = \sqrt{1 + m^2}$ yields
$z(x_t) = a\sqrt{1+m^2}$ and therefore
\begin{equation}
  h(x) = a \cosh\!\left(\frac{x - x_0}{a}\right) - a\sqrt{1 + m^2},
  \label{eq:height}
\end{equation}
which satisfies $h(x_t) = 0$ by construction. The surface height at the
target is the reference datum, and corners above or below that level
receive positive or negative heights accordingly.

\subsection{Sag Parameter and Minimum Fabric Length}

The sag parameter relates the fabric length to the module span. For a
catenary spanning a horizontal distance $D$, the arc length is
\begin{equation}
  S = 2a\sinh\!\left(\frac{D}{2a}\right).
  \label{eq:arc_length}
\end{equation}
When $S = D$ the sheet is taut and $a \rightarrow \infty$, producing a
straight line. As $S$ grows beyond $D$ the sheet develops slack, $a$
decreases and the profile sags more deeply, which widens the range of
achievable slopes but increases the actuator displacement needed to
produce them. For the platform of Section~\ref{sec:platform}, with
$S = 0.6$\,m and $D = 0.5$\,m, numerical solution
of~\eqref{eq:arc_length} gives $a = 0.235$\,m.

A geometric lower bound on the fabric length follows from the requirement
that the sheet must still connect adjacent actuators when they sit at
opposite extremes of their range. With one actuator raised to $+u_{\max}$
and its neighbour lowered to $-u_{\max}$, the material must bridge a
horizontal distance $D$ and a vertical distance $2u_{\max}$, so that
\begin{equation}
  S_{\min} = \sqrt{D^2 + 4\,u_{\max}^2}.
  \label{eq:smin}
\end{equation}
At $S = S_{\min}$ the full actuator range is available; further slack
reduces the fraction of actuator displacement converted into surface
slope. For $D = 0.5$\,m and $u_{\max} = 0.16$\,m,
\eqref{eq:smin} gives $S_{\min} = 0.594$\,m, so the fabric length of the
platform sits marginally above the bound.

\subsection{Separable Surface and Corner Heights}

The surface of one module is modelled as the sum of two independent
cross sections, one per axis (Fig.~\ref{fig:catenary}):
\begin{equation}
  z(x,y) = h(x;\, a_x, m_x, x_t) + h(y;\, a_y, m_y, y_t),
  \label{eq:separable}
\end{equation}
where $(x_t, y_t)$ is the target position. This decomposition matches the
four corner actuator layout, which provides two independent tilt axes. For
an actuator corner $j$ at position $\mathbf{c}_j = (c_j^x, c_j^y)$ the
desired height is $q_j = h(c_j^x) + h(c_j^y)$. The four heights share a
nonzero common offset that would translate the surface vertically without
altering its slope, so the mean is removed to devote the full stroke to
shaping:
\begin{equation}
  \bar{q}_j = q_j - \frac{1}{N}\sum_{k=1}^{N} q_k .
  \label{eq:mean_remove}
\end{equation}

%% file: sections/model_validation.tex
\section{Surface Model Validation}
For validation, the catenary is compared with a planar interpolant of
the support heights, as used by the baseline
controller~\cite{11163787}, the VVCM~\cite{hu2022multirobot} in the
stable kinematics formulation of~\cite{11353119} generalised to unequal support heights, and the isotropic membrane strain energy model of~\cite{HUNTE2021437}. The comparison covers surface shape, object equilibrium prediction and computational cost, and the catenary is further checked under object loading.

\subsection{Height and Slope Error}
\label{sec:validation}
\label{sec:val_setup}

\begin{figure*}[t]
  \centering
  \includegraphics[width=\textwidth]{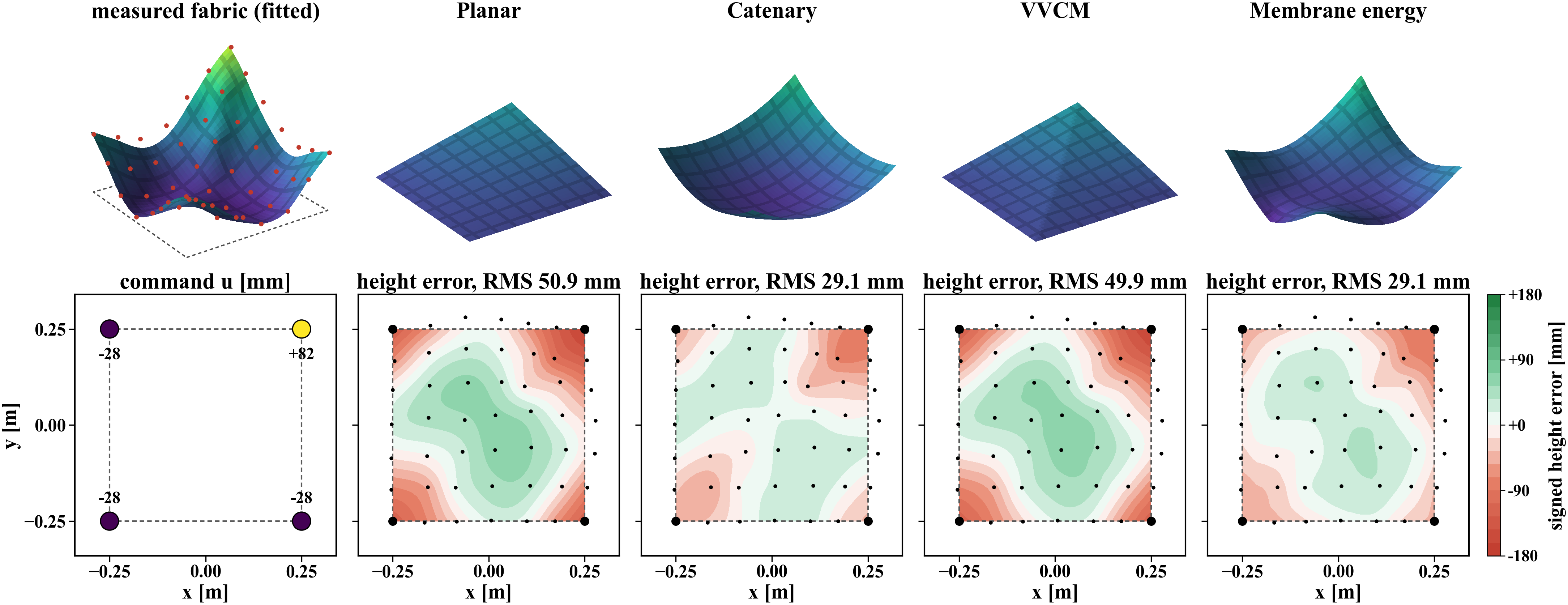}
  \caption{Surface models against the measured fabric for the command
  $[+82.5, -27.5, -27.5, -27.5]$\,mm. \textbf{Top:} the surface fitted
  through the 49 tracked markers (red dots) and the four model surfaces.
  \textbf{Bottom:} the applied command, then each model's signed height
  error over the support square; small dots mark the marker lattice,
  large dots the supports.}
  \label{fig:surface_hw}
\end{figure*}

Two surface references are used: simulation and hardware. The simulated reference
is the single 0.5 m pitch module MuJoCo membrane of Section~\ref{sec:platform}. The hardware-measured
reference is a $7 \times 7$ grid of retroreflective markers bonded at
0.1\,m spacing to the fabric of one module and recovered by motion
capture.  
A thin plate spline through the 49 tracked markers,
smoothed gives the measured surface, and every
model is scored against it on a $41 \times 41$ grid over the support
square, so the whole sheet is judged rather than the marker positions
alone. The height error $\epsilon_z$ is the RMS height difference after
the mean height is removed from both surfaces, the vertical datum of a
capture being arbitrary; the slope error $\epsilon_\nabla$ is the RMS
norm of the difference between predicted and measured surface
gradients. Three captures are taken on each reference with identical support commands: the command of Fig.~\ref{fig:surface_hw}, which raises one support 82.5\,mm above the mean and lowers the other three by 27.5\,mm, and two near the resting height. On hardware the marker grid extends beyond the 0.5\,m support square, and only the region inside the square is scored.


\label{sec:val_fidelity}
\label{sec:aconsistency}

\begin{table}[t]
  \centering
  \caption{Height error $\epsilon_z$ (mm) and slope error $\epsilon_\nabla$
   (dimensionless) of the four models. Top: the single command of
  Fig.~\ref{fig:surface_hw}, one RMS value per model. Bottom: mean
  $\pm$ sd over the three support configurations.}
  \label{tab:surface}
  \footnotesize
  \setlength{\tabcolsep}{3.5pt}
  \begin{tabular}{@{}lcccc@{}}
    \toprule
    & \multicolumn{2}{c}{Simulation} & \multicolumn{2}{c}{Hardware} \\
    \cmidrule(lr){2-3}\cmidrule(lr){4-5}
    Model & $\epsilon_z$ [mm] & $\epsilon_\nabla$ & $\epsilon_z$ [mm] & $\epsilon_\nabla$ \\
    \midrule
    \multicolumn{5}{@{}l}{\emph{Command of Fig.~\ref{fig:surface_hw}, $[+82.5, -27.5, -27.5, -27.5]$\,mm}} \\
    Planar          & 39.4 & 0.37 & 50.9 & 0.67 \\
    Catenary        & 15.9 & 0.23 & \textbf{29.1} & \textbf{0.41} \\
    VVCM            & 30.9 & 0.33 & 49.9 & 0.65 \\
    Membrane energy & \textbf{6.5} & \textbf{0.08} & \textbf{29.1} & 0.42 \\
    \midrule
    \multicolumn{5}{@{}l}{\emph{Three configurations, mean $\pm$ sd}} \\
    Planar          & $41.5 \pm 2.7$ & $0.40 \pm 0.03$ & $59.2 \pm 6.8$  & $0.63 \pm 0.03$ \\
    Catenary        & $15.4 \pm 0.5$ & $0.19 \pm 0.04$ & $\mathbf{43.6 \pm 12.4}$ & $\mathbf{0.44 \pm 0.08}$ \\
    VVCM            & $31.9 \pm 1.9$ & $0.32 \pm 0.03$ & $63.1 \pm 12.0$ & $0.65 \pm 0.09$ \\
    Membrane energy & $\mathbf{7.7 \pm 2.1}$ & $\mathbf{0.11 \pm 0.02}$ & $44.2 \pm 15.1$ & $0.47 \pm 0.09$ \\
    \bottomrule
  \end{tabular}
\end{table}

Table~\ref{tab:surface} and Fig.~\ref{fig:surface_hw} give the result.
On the slack physical sheet the catenary matches the best model, the
membrane energy model, on height error (29.1\,mm for both, a factor
1.7 below the planar interpolant and the VVCM) and has the lowest slope
error of the four, with one scalar and no numerical solution. 
On the simulated membrane the membrane energy model is more accurate
(6.5 against 15.9\,mm), the price of the catenary's simplicity, but the catenary is more accurate than the VVCM on both references and, as
Section~\ref{sec:val_cost} shows, far cheaper than either. The four
models rank the same on the simulated membrane as on the fabric, so a
result obtained in simulation in Section~\ref{sec:control} transfers
qualitatively to hardware.
All four models share one error pattern
(Fig.~\ref{fig:surface_hw}): too shallow at the centre and too low near
the supports, where the slack sheet, which continues beyond the support
square, runs straighter into the supports than any of the models.

\subsection{Equilibrium Prediction}
\label{sec:val_equilibrium}

\begin{table}[t]
  \centering
  \caption{Test objects and normalised equilibrium error
  $\bar{e} = e_p/D$ in \%, where $e_p$ is the mean distance in
  simulation between the settled object position and the minimum of
  each model's surface, over 40 corner configurations per object.}
  \label{tab:objects}
  \footnotesize
  \setlength{\tabcolsep}{3.5pt}
  \begin{tabular}{@{}llccc@{}}
    \toprule
    Object (mass, size in cm) & Motion & Catenary & VVCM & Membrane \\
    \midrule
    Sphere (32\,g, $\varnothing$4.5)             & rolling & \textbf{7.1}  & 8.3  & 18.4 \\
    Egg (61\,g, $\varnothing$4.1 to 6.2)         & rolling & \textbf{8.8}  & 14.6 & 25.5 \\
    Cube (31\,g, 4.2)                            & mixed   & \textbf{14.9} & 21.7 & 33.2 \\
    Cylinder (71\,g, $\varnothing$3 $\times$ 13) & mixed   & \textbf{12.1} & 15.6 & 25.3 \\
    Disk (26\,g, $\varnothing$7 $\times$ 2.5)    & sliding & \textbf{14.3} & 20.1 & 31.4 \\
    \bottomrule
  \end{tabular}
\end{table}

A second test is where an object comes to rest. The five objects of
Table~\ref{tab:objects} are placed at rest at random positions on the
simulated membrane under 40 random corner configurations each and left
to settle without control, and the settled position is compared with
the minimum of each model's surface; the planar interpolant has no
minimum and is excluded. The catenary is closest for every object,
within 7 to 15\% of the module span, and its error grows least from the sphere to the flat disk. The VVCM matches it for the sphere, the point mass its derivation assumes, and falls behind for every other shape. The membrane energy model, best on height, is worst here because its surface is nearly flat around the minimum, so the position of the minimum is poorly determined and a small height error moves it by several centimetres.

\subsection{Validity Under Object Loading}
\label{sec:loading}

To assess the sensitivity of the unloaded surface approximation to
object loading, the fabric is captured with all actuators at their
resting height, without an object, with a 50.3\,g object representative
of those manipulated in this paper, and with a 300\,g object as a
heavier stress case, each placed at the module centre or 0.17\,m from
it. At rest the actuators sit at their calibrated zero, and the sag
parameter fitted on six unloaded captures, $a = 0.248$\,m, is within
6\% of the geometric value; it is held fixed for the loaded captures.
The representative load sinks the sheet by 24\,mm at the centre and
raises the catenary height error from 12 to 17\,mm; the stress case
sinks it by 42\,mm and raises the error to 24\,mm. Refitting $a$ to
each loaded capture recovers at most 3\,mm of this error, so most of
the degradation is load induced deformation that changing the sag
parameter alone cannot represent. Placed off centre, either load lowers the sheet mostly near the centre, a global shift rather than a local pocket, and leaves the height error within 1.3\,mm of the unloaded value. For the objects of this paper, 26 to 71\,g, the unloaded catenary therefore holds to within 5\,mm of height error. Under a payload an order of magnitude heavier the sheet is no longer an exact catenary, but exact reconstruction of the loaded surface is not required for control: the model is expected to remain usable as an approximate local slope model for generating actuator commands.

\subsection{Computational Cost}
\label{sec:val_cost}

\begin{table}[t]
  \centering
  \caption{Cost of one forward evaluation per model on rigs of $N$
  supports. $T(N) = \sum_{K=3}^{N}\binom{N}{K}$
  is the number of VVCM taut sets (5, 466 and 65\,399); $M$ the number of
  mesh nodes (225, 441 and 625). Inverse: cost of the closed form map
  from commanded slope to support heights. $^{*}$Extrapolated from the
  measured per set cost.}
  \label{tab:cost}
  \footnotesize
  \setlength{\tabcolsep}{2.5pt}
  \begin{tabular}{@{}llrrrl@{}}
    \toprule
    & & \multicolumn{3}{c}{Latency [ms]} & \\
    \cmidrule(lr){3-5}
    Model & Complexity & $N{=}4$ & $N{=}9$ & $N{=}16$ & Inverse \\
    \midrule
    Planar & $O(N)$ & \textbf{0.02} & \textbf{0.04} & \textbf{0.06} & \textbf{20\,$\mu$s} \\
    Catenary & $O(N)$ & 6.9 & 9.3 & 9.8 & \textbf{20\,$\mu$s} \\
    VVCM, general $z_j$ & $O(T(N)\,\mathrm{poly}\,K)$ & 18.4 & 2767 & $6.1{\times}10^{5}$\,$^{*}$ & none \\
    VVCM, uniform $z_r$ & $O(T(N)\,K^3)$ & 0.17 & 25.5 & 3140 & none \\
    Membrane energy & $O(\mathrm{iters}\cdot M)$ & 240 & 488 & 675 & none \\
    \bottomrule
  \end{tabular}
\end{table}

Table~\ref{tab:cost} gives the cost of one forward evaluation of each model on rigs of $N = \{ 4, 9, 16 \}$ supports in a single threaded Python implementation; the scaling, not the absolute time, is the result. The planar and catenary models are linear in N, and their inverse, the closed form map ~\eqref{eq:vertex} to~\eqref{eq:mean_remove} from a commanded slope to the four support heights, costs under 20\,$\mu$s, which is what lets the model run inside the loop; the catenary forward time in the table is dominated by locating the surface minimum, which the controller never needs. The VVCM enumerates every taut set, so its cost grows as $2^N$, from 18\,ms at $N = 4$ to 2.8\,s
at $N = 9$, and the closed form of ~\cite{11353119} applies only to supports at one height. The membrane energy model minimises a finite element energy over the mesh at about 1.1\,ms per node. Neither has an inverse: the VVCM returns an equilibrium point rather than a slope field, and the membrane model defines slope only after minimisation.

%% file: sections/control.tex
\section{Catenary Position Controller}
\label{sec:control}

The model is validated in closed loop by replacing the geometric
transformation of the planar baseline~\cite{11163787}, which maps PID
tilt angles to actuator heights through the point normal equation of a
plane, with~\eqref{eq:height} to~\eqref{eq:mean_remove}. Control acts
on one module at a time through its four corner actuators at 60\,Hz;
Algorithm~\ref{alg:controller} gives one step.

\subsection{Controller}
\label{sec:controller}

\begin{algorithm}[t]
\caption{Catenary controller, one control step.}
\label{alg:controller}
\begin{algorithmic}[1]
\REQUIRE error $\mathbf{e}$, velocity $\mathbf{v}$, target $\mathbf{p}_t$,
  corners $\{\mathbf{c}_j\}_{j=1}^{4}$
\FOR{each axis $i \in \{x, y\}$}
  \STATE $I_i \leftarrow \mathrm{sat}_{I_{\max}}(I_i + e_i \Delta t)$,
         \quad $\dot{e}_i \leftarrow (e_i - e_i^{k-1})/\Delta t$
  \STATE $m_i^{\mathrm{raw}} \leftarrow -(K_p e_i + K_i I_i + K_d \dot{e}_i) + K_v v_i$
  \STATE $m_i \leftarrow \alpha\,\mathrm{sat}_{m_{\max}}(m_i^{\mathrm{raw}})
         + (1-\alpha)\, m_i^{k-1}$
\ENDFOR
\STATE $q_j \leftarrow h(c_j^x; a_x, m_x, p_{t,x}) + h(c_j^y; a_y, m_y, p_{t,y})
       \;\; \forall j$
\STATE $\bar{q}_j \leftarrow q_j - \mathrm{mean}(\mathbf{q}) \;\; \forall j$
\STATE $u_j \leftarrow \mathrm{sat}_{u_{\max}}(\bar{q}_j + \alpha_v \delta_j),
       \;\; \delta_j \sim \mathcal{U}(-1, 1) \;\; \forall j$
\end{algorithmic}
\end{algorithm}

A PID controller with velocity damping acts on each axis of the
position error and outputs a commanded slope (line 3), saturated and
low pass filtered (line 4) before~\eqref{eq:height}
and~\eqref{eq:mean_remove} convert it into corner heights (lines 6
and 7). The velocity term, fed by the tracking system, adds a counter
slope as the object approaches the target and limits overshoot near
edges and corners where objects gain speed; the vibration term
$\alpha_v \delta_j$ dislodges non spherical objects that settle into
fabric indentations and is $0.05$\,m for sliding and mixed objects and
zero for rolling ones. The gains, tuned by a genetic algorithm on the
integrated absolute error of target reaching trials in simulation, are
$K_p = 3.71$\,m$^{-1}$, $K_i = 1.81$\,m$^{-1}$s$^{-1}$,
$K_d = 0.86$\,s\,m$^{-1}$ and $K_v = 1.05$\,s\,m$^{-1}$, with
$I_{\max} = 1.0$\,m\,s, $m_{\max} = 1$, $\alpha = 0.3$ and
$a_x = a_y = 0.235$\,m from~\eqref{eq:arc_length}.

Because~\eqref{eq:separable} sums one function of $x$ and one of $y$,
the four corner heights of~\eqref{eq:mean_remove} are coplanar for every
$(m_x, m_y)$, as are those of the planar baseline: the two controllers
command the same family of corner planes and differ only in the map
from commanded slope to that plane. The planar map tilts the plane by
$m_i$ wherever the target lies; the catenary map tilts it by
$-2a\sinh(x_{0,i}/a)\sinh(D/2a)/D$ with $x_{0,i}$
from~\eqref{eq:vertex}, which is nonzero at $m_i = 0$ whenever the
target is off centre, placing the vertex of the sag at the target, and
whose sensitivity to $m_i$ grows with the distance of the vertex from
the centre. At $m_i = 0$ the planar map commands a level plane, under which an object at an off centre target rolls toward the sag minimum, whereas the catenary map commands the tilt that places the sag minimum at the target. The catenary controller is thus a position dependent feedforward tilt and gain schedule on the planar one, and no constant retuning of the planar gains reproduces it.


\subsection{Catenary Against Planar Control in Simulation}
\label{sec:sim_comparison}

\begin{figure}[t]
 \centering
 \includegraphics[width=\columnwidth]{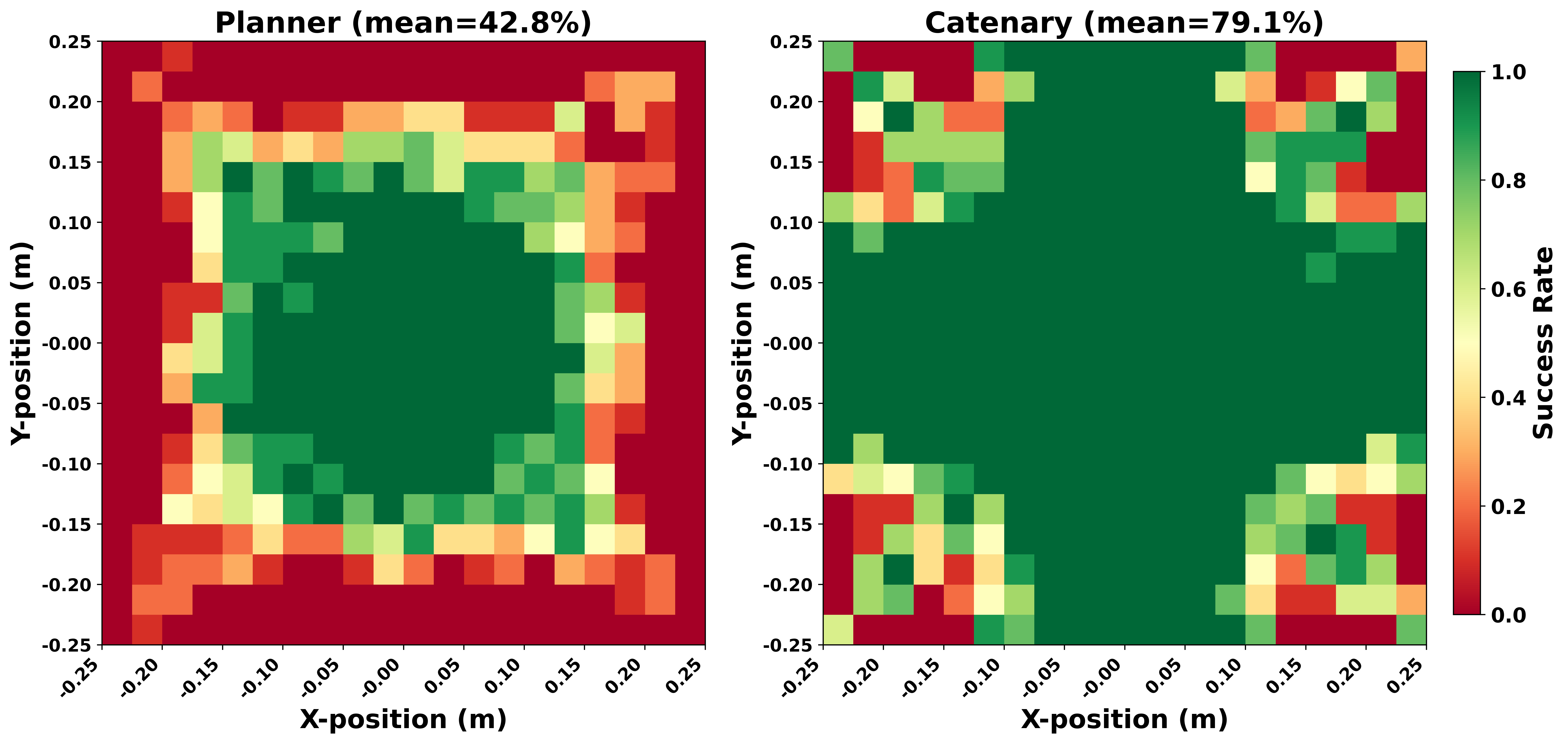}
 \caption{Success rate maps for a sphere on a single module in simulation.
 (a) Planar controller~\cite{11163787}. (b) Catenary controller.}
 \label{fig:heatmap_comparison}
\end{figure}

To measure the success rate in
simulation on a single module: the workspace is divided into 400
subsections of $0.025 \times 0.025$\,m, each receives 10 trials with a
random target inside it and a random initial object position, and a
trial succeeds if the object reaches within $\rho = 0.03$\,m of the
target inside $t_{\max} = 10$\,s without leaving the surface. Fig.~\ref{fig:heatmap_comparison} gives the maps for the sphere. The
planar controller attains a mean success rate of 42.8\%, with success
confined to a scattered band through the centre; the catenary
controller attains 79.1\%, and its success region is one contiguous
area covering most of the module interior. Residual failures at the corners are objects leaving the
module edge, which the simulated contact does not fully prevent. The gains are largest in the central and near edge regions, where the feedforward tilt of Section~\ref{sec:controller} holds an off centre target that the planar controller cannot.

%% file: sections/design.tex
\section{Design of Suspended Fabric Surfaces}
\label{sec:design}

\begin{figure*}[t]
  \centering
  \includegraphics[width=\textwidth]{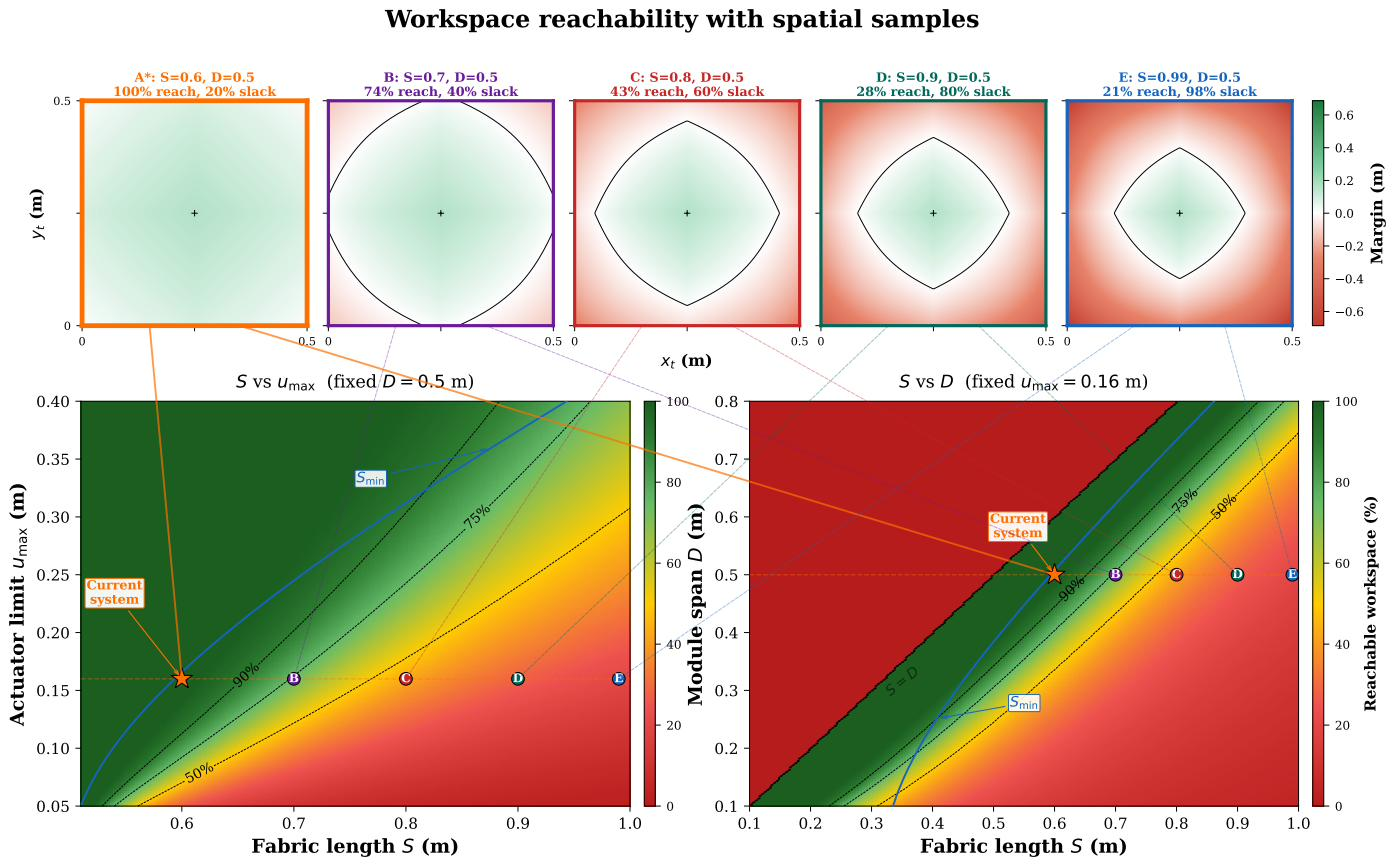}
  \caption{Design space of a suspended fabric module. \textbf{Top:} margin maps
  $\mu(x_t, y_t)$ of~\eqref{eq:margin} at five operating points A$^{*}$ to
  E of increasing fabric slack; green is positive margin, red is actuator
  saturation and the black contour is the zero margin boundary.
  \textbf{Bottom:} reachable workspace percentage over fabric length and
  actuator limit (left) and over fabric length and module span (right),
  with the blue curve giving $S_{\min}$ from~\eqref{eq:smin} and the star
  marking the operating point of the platform. Parameters not varied are
  held at $D = 0.5$\,m and $u_{\max} = 0.16$\,m.}
  \label{fig:reachability}
\end{figure*}

Of the four surface models, only the catenary is both expressed in
the quantities a designer selects, the module span $D$, the fabric
length $S$ and the actuator limit $u_{\max}$, and invertible from a
commanded slope to support heights (Section~\ref{sec:val_cost}), which
is the map a reachability analysis evaluates at every candidate target.
This section derives from it the fraction of a module that the
controller can hold as a function of those three quantities.

\subsection{Reachability Condition}
\label{sec:reachability}

A target $\mathbf{p}_t = (x_t, y_t)$ is defined as reachable if the
controller can hold a zero corrective slope at $\mathbf{p}_t$ without any
actuator exceeding its limit, that is if the configuration
$m_x = m_y = 0$ with $\alpha_v = 0$ in algorithm~\ref{alg:controller} satisfies
$|u_j| \leq u_{\max}$ for all $j$. The critical case arises when the
target lies near a module corner, where the profile must produce a large
height difference between the near and far actuators. The condition is
evaluated pointwise across a module. For each candidate target the margin
is the minimum remaining actuator headroom,
\begin{equation}
  \mu(x_t, y_t) = u_{\max} - \max_{j} |u_j| ,
  \label{eq:margin}
\end{equation}
and the fraction of module area with $\mu > 0$ defines the reachable
workspace percentage. The condition tightens to a required slope reserve $m_r$ by
evaluating~\eqref{eq:margin} at $|m_x| = |m_y| = m_r$; the charts below
use $m_r = 0$, the weakest form.

\subsection{Design Charts}

Fig.~\ref{fig:reachability} presents the resulting design space, with
slack defined as $(S - D)/D$. At the operating point of the platform,
$S = 0.6$\,m and 20\% slack, the reachable region covers the whole
module; as the fabric lengthens it contracts from the corners inward, to
74\% at $S = 0.7$\,m and 21\% at $S = 0.99$\,m, where it is confined
to a small diamond about the centre. In the lower left panel the
$S_{\min}$ curve of~\eqref{eq:smin} separates sheets too short to
connect actuators at opposite extremes of their range from sheets whose
slack costs coverage; the lower right panel shows that larger spans
require proportionally more material for the same coverage.

The charts turn the selection of platform parameters into a reading of
contours: given a span fixed by the application and a required
coverage, the admissible band of fabric length lies between the
$S_{\min}$ curve and the contour of that coverage, and its width is the
tolerance to fabric stretch and wear. For the present platform that band
is narrow: the operating point lies 6\,mm above $S_{\min}$, and a
0.1\,m increase in fabric length would remove the corners from the
reachable set unless the actuator limit were raised accordingly.

%% file: sections/multimodular.tex
\section{Multimodular Extension}
\label{sec:multimodular}


To extend the workspace beyond a single module, actuators are arranged on
an $(n+1) \times (n+1)$ grid forming $n \times n$ modules that share
boundary actuators with their neighbours. In the $2 \times 2$
configuration of Fig.~\ref{fig:platform}B, nine actuators define four
modules M0 to M3 covering $1m^2$. Adjacent modules share two
boundary actuators and diagonal modules share one, so the actuator density
falls from sixteen to nine, a reduction of 44\% at this scale that
approaches 75\% asymptotically. Each actuator operates as an independent
unit, which preserves robustness to individual faults.


Objects are transferred by raising the actuators that belong exclusively
to the current module while lowering the remaining actuators, namely the
two shared with the destination module and the two belonging to it. The
resulting local slope induces rolling, sliding or a combination of the
two. Transfer operates hierarchically: a planner based on Manhattan
distance determines the module sequence and enforces conflict free paths,
so that objects under simultaneous manipulation never occupy the same
module, while a lower layer executes each transfer as a timed actuator
sequence with feedback from the tracking system and adaptive retries at
increased actuator height and hold duration. The lateral path of the
transfer is selected through the choice of shared actuator to raise:
raising the inner shared actuator produces a left path, raising the outer
one produces a right path, and raising both equally produces a central
path. Once the object rests on the destination module, the catenary
controller of Section~\ref{sec:control} acts on that module alone.

\subsection{Target Reaching on multi-modular Platform}
\label{sec:target}
\begin{table}[t]
  \centering
  \caption{End to end target reaching with the catenary controller on
  the $2 \times 2$ simulated membrane and hardware platform: two
  trials per object. SR$_3$: trials that came within 3\,cm of the
  target at any time; $t_3$: time from trial start, transfer included,
  to first reach 3\,cm, over successful trials; $e_{\mathrm{fin}}$:
  final distance to the target. Mean $\pm$ sd. Every trial came within
  5\,cm except one hardware cylinder trial.}
  \label{tab:hw_target}
  \footnotesize
  \setlength{\tabcolsep}{3pt}
  \begin{tabular}{@{}lcccccc@{}}
    \toprule
    & \multicolumn{3}{c}{Simulation} & \multicolumn{3}{c}{Hardware} \\
    \cmidrule(lr){2-4}\cmidrule(lr){5-7}
    Object & SR$_3$ & $t_3$ [s] & $e_{\mathrm{fin}}$ [cm]
           & SR$_3$ & $t_3$ [s] & $e_{\mathrm{fin}}$ [cm] \\
    \midrule
    Sphere   & 2/2 & $6.2 \pm 1.0$  & $0.9 \pm 0.0$ & 2/2 & $6.4 \pm 0.4$  & $1.6 \pm 0.7$ \\
    Cube     & 2/2 & $16.7 \pm 1.2$ & $1.9 \pm 0.3$ & 2/2 & $8.5 \pm 2.8$  & $1.9 \pm 0.4$ \\
    Disk     & 2/2 & $19.2 \pm 0.5$ & $2.7 \pm 0.2$ & 2/2 & $11.1 \pm 2.5$ & $2.3 \pm 0.1$ \\
    Egg      & 2/2 & $8.1 \pm 0.1$  & $1.6 \pm 0.3$ & 1/2 & 5.3            & $2.1 \pm 2.5$ \\
    Cylinder & 2/2 & $13.5 \pm 0.9$ & $1.5 \pm 0.5$ & 1/2 & 9.9            & $3.7 \pm 3.2$ \\
    \bottomrule
  \end{tabular}
\end{table}
 
Target reaching evaluates end to end manipulation with the five objects
of Table~\ref{tab:objects}: transfer along a Manhattan path to the
target module, then the catenary controller on that module, on the
$2 \times 2$ membrane of Section~\ref{sec:platform} and on the hardware
platform. Table~\ref{tab:hw_target} reports the final error
per object for two runs. In simulation every object comes within 3\,cm of the target in both trials and finishes 0.9 to 2.7\,cm from it; on hardware the sphere, cube and disk do the same in 6 to 11\,s including transfer, while the egg and cylinder reach 3\,cm in one of two trials each, the cylinder with the largest final error. Rolling objects converge fastest on both references; the sliding cube and disk settle about twice as slowly in simulation, where the contact friction is lower than on the fabric.


\subsection{Interference From Shared Boundary Actuation}
\label{sec:interference}

\begin{table}[t]
  \centering
  \caption{Displacement of a stationary object on a passive module while
  an object is manipulated on a neighbouring module, in metres, mean of
  three repetitions. $d_{\mathrm{rms}}$, $d_{\max}$: root mean square
  and maximum displacement over the manipulation; $\Delta$: reduction of
  $d_{\mathrm{rms}}$ with compensation; lost: object left the module;
  NA: not tested.}
  \label{tab:interference}
  \footnotesize
  \setlength{\tabcolsep}{3pt}
  \begin{tabular}{llccccc}
    \toprule
     & & \multicolumn{2}{c}{No comp.}
       & \multicolumn{2}{c}{Comp.} & \\
    \cmidrule(lr){3-4}\cmidrule(lr){5-6}
    Scenario & Object & $d_{\mathrm{rms}}$ & $d_{\max}$ & $d_{\mathrm{rms}}$ & $d_{\max}$
             & $\Delta$ \\
    \midrule
    Target, adjacent & Sphere & 0.217 & 0.460 & 0.047 & 0.078 & 78\% \\
    Target, adjacent & Disk   & 0.064 & 0.076 & 0.026 & 0.040 & 59\% \\
    Target, diagonal & Sphere & 0.031 & 0.058 & NA    & NA    & NA   \\
    Target, diagonal & Disk   & 0.035 & 0.044 & NA    & NA    & NA   \\
    Transfer, perp.& Sphere & 0.044 & 0.065 & 0.031 & 0.044 & 30\% \\
    Transfer, perp.& Disk   & 0.057 & 0.148 & 0.022 & 0.036 & 61\% \\
    Transfer, opp.& Sphere & lost  & lost  & 0.015 & 0.030 & NA   \\
    Transfer, opp.& Disk   & lost  & lost  & 0.009 & 0.018 & NA   \\
    \bottomrule
  \end{tabular}
\end{table}

A shared actuator serves two modules, so manipulating an object on one
module deforms the surface of its neighbours. The perturbation is
quantified by tracking a stationary object on a passive module while
another object is manipulated on a neighbouring one, and is reported
through the root mean square displacement over the manipulation duration
and the maximum displacement. A sphere and a disk serve as stationary
objects, representing rolling dominant and sliding dominant geometries.
Four spatial configurations are evaluated: target reaching on a module
that shares two actuators with the passive module (adjacent) or one
(diagonal), and transfer that activates one shared actuator
(perpendicular) or both (opposite).

Table~\ref{tab:interference} summarises the results. Without
compensation, the adjacent configuration produces severe displacement of
the sphere, with a root mean square value of 0.217\,m and a maximum
displacement of 0.460\,m, comparable with the module span, indicating
exit from the module. The disk shows lower but substantial sensitivity,
its flat contact geometry resisting rolling. The diagonal configuration
couples markedly less, consistent with interference scaling with the
number of shared actuators. During transfer with both shared actuators
raised, the stationary object is displaced beyond the module boundary and
is lost.

The simple first-step compensation raises the two actuators of the passive module
that are not shared with the active module. When the shared actuators move
during manipulation on the active module, the passive surface develops an
unintended tilt toward it; raising the non shared actuators creates a
counter tilt that restores a locally level surface. Compensation reduces
the root mean square displacement by 30 to 78\% across
configurations, brings the adjacent configuration to a level comparable
with the uncompensated diagonal configuration, and prevents object loss
during transfer with both shared actuators active.

%% file: sections/discussion.tex
\section{Discussion}
\label{sec:discussion}

\textbf{Model.} Fabric elasticity and wear are not represented, the sag
parameter is set offline, and the unloaded assumption restricts the
model to light objects (Section~\ref{sec:loading}). The hardware
ordering of the surface models rests on three captures and is resolved
on the matched command but only partially in the held out aggregate, and
the equilibrium comparison is simulation only. The controller comparison
of Section~\ref{sec:sim_comparison} is performed on the tensioned
simulated membrane, so its success rates characterise the two
controllers relative to each other rather than predicting hardware
coverage, and the gains were tuned through the catenary map, so part of
the margin may reflect tuning.

\textbf{Design.} The reachability condition is geometric. It does not
account for the friction and object mass that motion must overcome, so
a target reachable in the charts may be unreachable for a heavy or high
friction object, nor for pulling, in which fabric tension displaces the
object without a net slope. The charts use the geometric sag parameter
$a = 0.235$\,m, whereas the command driven calibration of
Section~\ref{sec:aconsistency} returns 0.335\,m in simulation and
0.440\,m on hardware; a larger $a$ flattens the profile and widens the
reachable set, so the charts are conservative with respect to the
calibrated values, and they have been checked against the platform at
one operating point.

\textbf{Multimodular platform.} Transfer between modules relies on an open
loop actuator sequence whose outcome is checked only after the hold, so
a failed transfer costs a full retry; the planner admits at most one actively manipulated
object per module, and shared boundary actuation is not compared against
independent actuation of the same grid. The platform is demonstrated at
the $2 \times 2$ scale on hardware only; in a $3 \times 3$ grid the
central module shares actuators with four neighbours, so the growth of
interference sources is expected to require decentralised compensation.
Object position is tracked by an external motion capture system, which
bounds deployability and is a single point of failure. 
 
 
Future work includes extending the model with a load term that depends
on the mass and position of one or several objects, which would cover
heavier payloads, estimating the sag parameter online to follow fabric
wear, and adding friction and object mass to the reachability condition
to tighten the charts. Tracking the object during the hardware captures
would allow the equilibrium test of Section~\ref{sec:val_equilibrium}
to be repeated on the physical fabric, and repeating the success rate
maps of Section~\ref{sec:sim_comparison} on hardware would confirm the
controller comparison with real contact and friction.

%% file: sections/conclusion.tex
\section{Conclusion}
\label{sec:conclusion}

This paper presented a separable catenary model of a suspended fabric
manipulation surface, a design space derived from it, and a multimodular
platform that extends the module to larger areas. On a matched command
the catenary ranks second of four surface descriptions by height error
on both a simulated membrane and the physical fabric, behind a membrane
energy model that cannot be evaluated at the control rate, and it
predicts the rest position of five object geometries better than any of
them; a controller built on it nearly doubles the mean target reaching
success rate of the planar baseline in simulation. Because the model is
expressed in module span, fabric length and actuator stroke, it yields a
minimum fabric length bound and charts of reachable workspace from which
platform parameters can be read directly. Shared boundary actuation
requires $(n+1)^2$ actuators for an $n \times n$ grid; on a $2 \times 2$
prototype heterogeneous objects transferred between modules reach
targets within 0.03\,m, and the interference the shared actuators
introduce is reduced by 30 to 78\% through actuator compensation.